%% file: main_final.tex
\documentclass[10pt,twocolumn,letterpaper]{article}

\usepackage[pagenumbers]{cvpr} %
\usepackage{multirow}
\usepackage{makecell}
\usepackage{overpic}
\usepackage{xstring}
\usepackage{pgffor}
\usepackage{xparse}
\usepackage{etoolbox}
\usepackage{graphicx}
\usepackage{subcaption}
\usepackage[table]{xcolor}
\input{utils.tex}

\input{preamble}
\definecolor{cvprblue}{rgb}{0.21,0.49,0.74}
\usepackage[pagebackref,breaklinks,colorlinks,allcolors=cvprblue]{hyperref}

\def\paperID{****} %
\def\confName{CVPR}
\def\confYear{2026}

\newcommand{\figref}[1]{Fig.~\ref{#1}}
\newcommand{\tabref}[1]{Table~\ref{#1}}
\newcommand{\secref}[1]{Sec.~\ref{#1}}

\newcommand{\para}[1]{\vspace{0.025in}\noindent\textbf{#1}\quad}

\title{RefOnce: Distilling References into a Prototype Memory \\ for Referring Camouflaged Object Detection}

\author{Yu-Huan Wu$^{1, 2}$\quad Zi-Xuan Zhu$^{1}$\quad Yan Wang$^{2}$\quad Liangli Zhen$^{2}$ \quad Deng-Ping Fan$^{1\dagger}$ \\
$^1$VCIP, CS, Nankai University\quad $^2$IHPC, A*STAR, Singapore \\
\texttt{\small wyh.nku@gmail.com}\quad \small{Project Page: \url{https://github.com/yuhuan-wu/RefOnce}}
}

\begin{document}
\maketitle

\footnotetext[2]{Corresponding Author: Deng-Ping Fan (fdp@nankai.edu.cn)}

\begin{abstract}
Referring Camouflaged Object Detection (Ref‑COD) segments specified camouflaged objects in a scene by leveraging a small set of referring images. 
Though effective,
current systems adopt a dual‑branch design that requires reference images at test time, which limits deployability and adds latency and data‑collection burden. 
We introduce a Ref‑COD framework that distills references into a class‑prototype memory during training and synthesizes a reference vector at inference via a query‑conditioned mixture of prototypes. 
Concretely, we maintain an EMA‑updated prototype per category and predict mixture weights from the query to produce a guidance vector without any test‑time references. 
To bridge the representation gap between reference statistics and camouflaged query features, we propose a bidirectional attention alignment module that adapts both the query features and  the class representation. 
Thus, our approach yields a simple, efficient path to Ref‑COD without mandatory references. 
We evaluate the proposed method on the large-scale R2C7K benchmark. 
Extensive experiments demonstrate competitive or superior performance of the proposed method compared with recent state-of-the-arts.
\end{abstract}

\section{Introduction}
\label{sec:intro}

Camouflaged object detection (COD) aims to segment targets that intentionally resemble their surroundings, which is a capability vital to many applications like medical imaging \cite{fan2020pranet, wu2021jcs, fan2020inf}, industrial inspection \cite{chu2010camouflage}, and crop pest monitoring~\cite{liu2019pestnet}.
Despite recent impressive advances \cite{fan2020camouflaged, fan2021concealed, pang2024zoomnext}, COD remains extremely challenging because camouflaged targets have very low visual contrast and their appearance cues are easily suppressed by cluttered backgrounds \cite{lv2023towards}. 
Intuitively, if we could observe the same category in cleaner, more salient conditions, such as reference images with distinct targets, 
the model could learn more stable, class-level semantics and transfer them to ambiguous camouflaged scenes. 
This simple intuition naturally leads to Referring Camouflaged Object Detection (Ref-COD) \cite{zhang2025referring}, an intriguing and emerging task where a small set of reference images with salient targets guides the detection of the specified camouflaged object.

\begin{figure}[t]
  \centering
  \includegraphics[width=\linewidth]{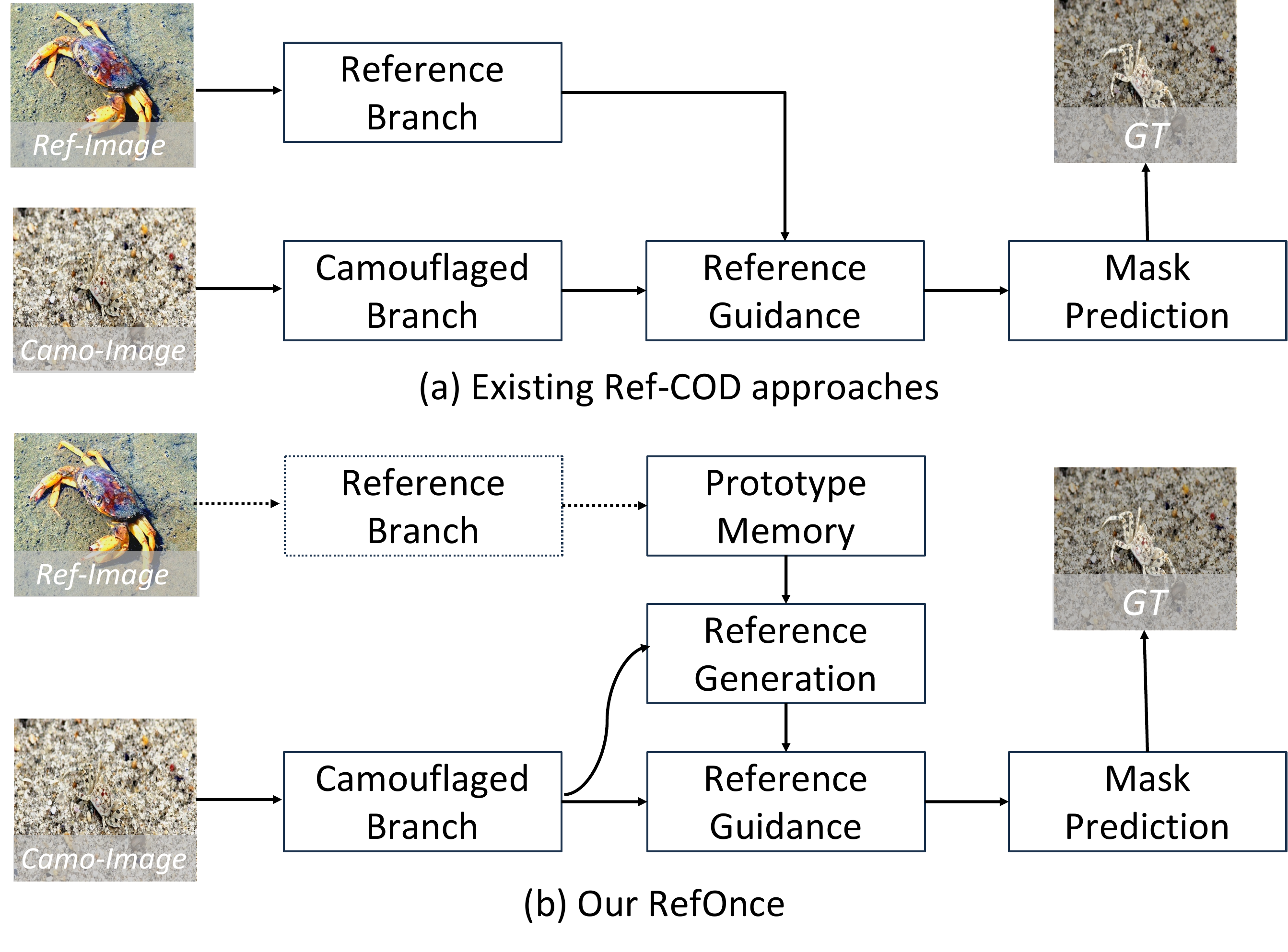}
  \vspace{-7mm}
  \caption{\textbf{Comparison between (a) existing Ref-COD approaches \cite{zhang2025referring, wu2025uncertainty} and (b) our RefOnce framework}. RefOnce retrieves the reference from the prototype memory to enable reference-free inference.}
  \label{fig:banner}
\end{figure}

Recent pioneers of Ref-COD \cite{zhang2025referring, wu2025uncertainty} have demonstrated that introducing reference images with clear backgrounds significantly improves the discriminability of COD models. 
R2CNet \cite{zhang2025referring} employs a dual-branch design: one branch extracts common representations from the reference images, while the other performs camouflaged segmentation under their modulation. Such designs successfully transform COD from blind searching into target-oriented detection, yielding strong quantitative gains and visually coherent masks. More advanced models like uncertainty-aware transformers \cite{wu2025uncertainty} further enhance the guidance to camouflaged object features from references. 

However, despite these successes, two core challenges remain. 
First, existing Ref-COD methods rely on a set of reference images of the same categories during inference to provide class-level guidance. 
However, in realistic deployment, this dependency is impractical because the camouflaged targets are visually ambiguous and category identification itself demands human efforts or external annotations. 
In other words, before the model can even segment the camouflaged object, one must first know its category to retrieve the corresponding reference images, a process that contradicts the goal of automatic detection. 
This challenge is further exacerbated by the scarcity of high-quality reference data and the privacy constraints in some application domains.
Second, there exists a significant gap between the salient reference domain and the low-contrast camouflaged query domain, causing weak spatial alignment and degraded guidance transfer.
Existing approaches mainly apply global affine modulation from reference features, which only weakly aligns spatial semantics and often causes over-transfer or misalignment. 
These issues motivate us to rethink the Ref-COD task: Can we distill the essential reference knowledge during training, enabling reference-free inference while simultaneously bridging the domain gap between reference and camouflaged features?

To address these challenges, we propose a novel Ref-COD framework \textbf{RefOnce}, which only relies on reference images during training. RefOnce is built upon two complementary ideas. 
First, to resolve the core deployment challenge, we design a class-prototype memory that stores category-level prototypes distilled from reference features during training, updated with an exponential moving average.
During inference, a lightweight predictor estimates query-conditioned mixture weights to synthesize a guidance vector entirely from these stored prototypes, eliminating the need for reference input. 
Second, to bridge the domain gap and effectively apply this synthesized guidance, we introduce a bidirectional alignment module that performs coupled attention between the synthesized guidance and query features to obtain a spatial gate, 
then applies the spatial modulation and a class-token offset to adapt both the query features and the class representation. 
Integrated into a modern COD backbone, this design provides flexible deployment and robust feature alignment.

Extensive experiments on the large-scale R2C7K benchmark and four standard COD datasets demonstrate the effectiveness and generalization of our framework. 
Under the standard Ref-COD protocol, our model achieves new state-of-the-art results across both CNN and Transformer backbones, even without test-time references. 
These results validate that our design preserves the semantic benefits of reference guidance while removing its practical constraints.

To summarize, our contributions are three-fold:

\begin{itemize}
    \item We address the core deployment paradox of Ref-COD. We propose \textbf{RefOnce}, the first framework to achieve fully reference-free inference, eliminating the impractical test-time dependency on reference images that hindered prior real-world application.
    
    \item To enable this paradigm, we design an efficient system that distills reference knowledge into a class-prototype memory and synthesizes adaptive guidance via a query-conditioned mixture mechanism. This system is complemented by a bidirectional attention alignment (BAA) module to effectively bridge the salient-to-camouflage domain gap.
    
    \item Extensive experiments demonstrate that RefOnce, despite requiring no test-time references, achieves new state-of-the-art performance on the large-scale R2C7K benchmark. Crucially, our design provides strong generalization to unseen categories, a capability fundamentally unavailable to previous reference-dependent methods.
\end{itemize}

\section{Related Work}
\label{sec:related}

\subsection{Camouflaged Object Detection}

Camouflaged Object Detection (COD) targets the segmentation of objects that deliberately mimic their surroundings, where extremely low contrast, background clutter, and subtle boundaries jointly impede reliable detection.
COD has accelerated with standardized datasets/baselines that consolidated evaluation and training practice \cite{fan2020camouflaged,fan2021concealed,lv2023towards,pang2024zoomnext,lv2021simultaneously,zhang2022preynet}. 
Architecturally, transformer-driven and multi-scale designs enhance global-to-local reasoning and fine details, such as CamoFormer \cite{yin2024camoformer}, ZoomNe(X)ts \cite{pang2022zoom,pang2024zoomnext}, Feature-Shrinkage Pyramid \cite{huang2023feature}, MSCAF \cite{liu2023mscaf}, and FAPNet \cite{zhou2022feature}. 
Boundary- and structure-aware modeling improves edge fidelity via explicit boundary decoders and reconstruction \cite{ji2022fast,zhu2022can,sun2022boundary,he2023feder,ye2025escnet}, complemented by texture/structure enhancement \cite{zhu2021inferring,chen2022boundary,li2022findnet,zhang2022preynet}.
Robustness is further advanced by uncertainty reasoning \cite{yang2021uncertainty,li2021uncertainty,liu2022modeling}, iterative refinement \cite{hu2023high,jia2022segment}, and graph interaction~\cite{zhai2021mutual,zhai2022mgl} to focus on ambiguous regions and hard pixels. 
Frequency- and texture-domain cues provide complementary signals for separating foreground from clutter (e.g., \cite{zhong2022detecting,lin2023frequency,sun2024frequency,zhang2025frequency, he2023feder}).
Depth cues further provide geometric priors that help distinguish camouflaged objects from visually similar backgrounds by revealing subtle spatial discontinuities~\cite{wu2023source,wang2023depth,wang2024depth}.
More works can be found in recent surveys \cite{fan2023advances, xiao2024survey, zhao2025deep}.

Despite these advances, COD still struggles with extremely low contrast, tiny targets, and cross-domain generalization \cite{fan2021concealed,lv2023towards}.
These limitations motivate Referring COD, which introduces explicit category-conditioned guidance from a few reference images to reduce ambiguity in low-contrast and small-object cases (e.g., \cite{zhang2025referring,wu2025uncertainty}).

\begin{figure*}[t]
  \centering
  \includegraphics[width=\linewidth]{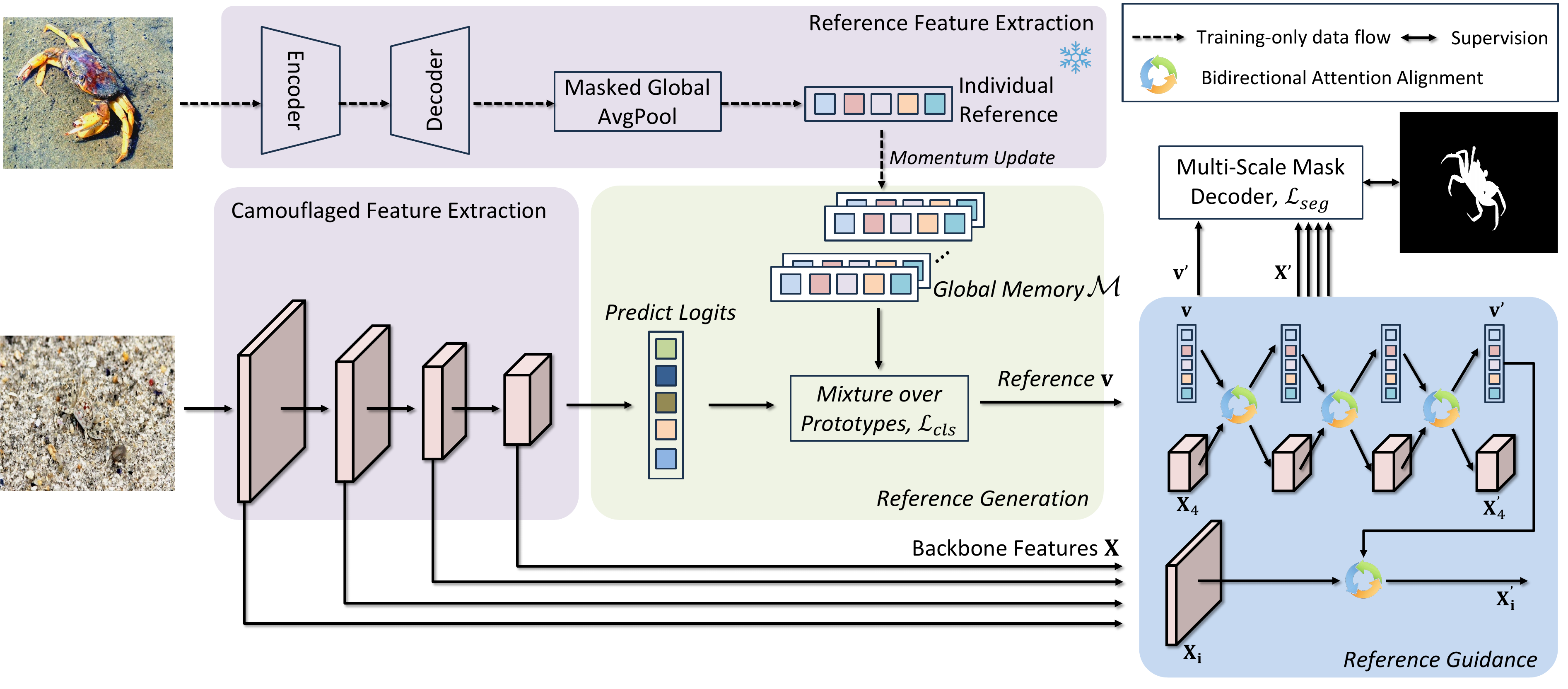}
  \vspace{-6mm}
  \caption{\textbf{Pipeline of the proposed RefOnce framework.}
  During training, the reference branch (top, dashed) distills features into a Global Memory $\mathcal{M}$; at inference, the reference $\mathbf{v}$ is synthesized from $\mathcal{M}$ and predict logits.}
  \label{fig:overview}
\end{figure*}

\subsection{Referring Camouflaged Object Detection}

Referring Camouflaged Object Detection (Ref-COD) aims to segment the camouflaged target specified by a few reference images from the same category, leveraging the reference images as category-consistent guidance for the query image. 
This task is motivated by the need to inject stable class-level cues for reliable detection in low-contrast and cluttered scenes. 
The seminal work R2CNet \cite{zhang2025referring} introduced a simple solution which employs a dual-branch design to extract the high-level target features from reference images and guide the camouflaged branch to recognize the camouflaged objects.
Subsequently, UAT \cite{wu2025uncertainty} proposed an uncertainty-aware transformer that integrates reference semantics into camouflaged feature learning via a cross-attention encoder and models token-level dependencies as Gaussian random variables through a probabilistic decoder.
RPMA \cite{liu2024reference} introduced an adapter that injects the reference cues to the transformer backbone with reference-generated dynamic convolutions.

However, these approaches still face two key limitations.
First, they require reference images during inference, which restricts practical deployment since obtaining suitable references for each query is costly and often infeasible.
Second, the salient-camouflage domain gap is insufficiently addressed, as global modulation or cross-attention from reference features often causes misalignment or over-transfer.
To overcome these issues, we propose RefOnce, which distills reference knowledge into a class-prototype memory during training and performs reference-free inference with a bidirectional alignment module for robust feature adaptation.

\section{Methodology}
\label{sec:method}

In this section, we will first summarize the overall architecture in \secref{sec:overview}.
Then, we will introduce the class-prototype memory and the bidirectional alignment in \secref{sec:memory} and \secref{sec:alignment}, respectively.

\subsection{Architecture}
\label{sec:overview}

\para{Preliminary.}
Ref-COD is an emerging task, which detects and segments the camouflaged objects within a natural image via the specific guidance of the reference images with clear targets in the same categories.
Given the input camouflaged image $I$ and the reference image $I_{ref}$ from the same category, the Ref-COD framework processes them separately via camouflaged and reference branches, and derives the multi-scale query features $\{\mathbf{X}_i\}$ and a reference vector $\mathbf{r}$.
With a guidance vector $\mathbf{v}$ obtained from $\mathbf{r}$ during training and synthesized from the prototype memory $\mathcal{M}$ at inference, the query branch performs reference guidance on $\{\mathbf{X}_i\}$, and the mask decoder predicts the binary mask $M$.

\para{Feature extraction.}
Our method follows the above basic framework.
The overall architecture of our method is illustrated in \figref{fig:overview}.
It consists of five parts: reference feature generation, camouflage feature extraction, class-prototype memory generation, reference guidance, and multi-scale mask decoder. For convenience, we take ResNet-50 backbone \cite{he2016deep} as the example.
Given the camouflaged image $I$, the backbone network will extract it to four-stage features as $\mathbf{X}_1$, $\mathbf{X}_2$, $\mathbf{X}_3$, and $\mathbf{X}_4$, with strides of 4, 8, 16, 32, respectively.
For the reference image $I_{ref}$, we employ the frozen ICON \cite{zhuge2022salient} network to extract the final saliency prediction features, followed by a multiplication of the foreground mask of the reference target. Then we conduct a global average pooling over each channel to derive the 1-D reference vector $\mathbf{r}$.
We then perform reference guidance on the multi-stage features $\mathbf{X}_i$ ($i\in[1,4]$) together with a guidance vector $\mathbf{v}$; in our reference-free instantiation, $\mathbf{v}$ is synthesized from the prototype memory at inference.

\para{Reference generation and guidance.}
We decouple learning from inference: during training, the model summarizes category cues from references into a compact prototype memory $\mathcal{M}$; at inference, given only a camouflaged query, a lightweight predictor consults this memory to produce a query-conditioned guidance vector $\mathbf{v}$, so no reference image or branch is required at test time. To make the guidance effective in low-contrast, cluttered scenes, we further apply a bidirectional alignment module that jointly adapts the query features and the guidance representation, and inject the guidance across multiple feature scales to highlight object-related regions while suppressing distractors. The specific designs of the prototype memory and the alignment will be detailed in \secref{sec:memory} and \secref{sec:alignment}.

\para{Mask prediction.}
Given the aligned multi-scale features $\{\mathbf{X}_i'\}$, we use a standard multi-scale mask decoder to predict the segmentation. In the regular instantiation, we first obtain a relevance seed by correlating the guidance vector with the top-stage refined feature map, and then inject this seed into a top-down mask decoder with progressive upsampling and lateral aggregation. 
For simplicity, we apply the mask decoder of ZoomNet \cite{pang2022zoom} as our basic mask decoder.

\para{Loss function.}
We supervise the model with two terms: (1) a binary cross-entropy (BCE) loss $\mathcal{L}_{\text{seg}}$ on the predicted segmentation masks; (2) a token/classification loss $\mathcal{L}_{\text{cls}}$ (\secref{sec:memory}) that supervises the mixture predictor for synthesizing $\mathbf{v}$ from the memory. The final objective is
$
\mathcal{L}=\mathcal{L}_{\text{seg}}+\lambda_c\,\mathcal{L}_{\text{cls}}
$
, where $\lambda_c{=}0.03$ in implementation.

\subsection{Reference Generation}
\label{sec:memory}

Prior Ref-COD frameworks \cite{zhang2025referring, wu2025uncertainty} adopt a reference-conditioned dual-branch design at test time, which leads to a core limitation: 
the dependency on reference images during inference. 
Fetching and encoding references increases latency and causes unstable guidance, as model behavior depends on which specific reference set is available. 
To eliminate this test-time dependency, we distill the category-level semantics of references into a prototype memory during training, 
which can be efficiently queried at inference to achieve reference-free, category-aware detection.

Let $\mathcal{M}=\{\mathbf{m}_1,\dots,\mathbf{m}_K\}$ be a learnable prototype memory (where $K$ is the number of prototypes, typically the number of categories). Each prototype $\mathbf{m}_k\in\mathbb{R}^C$ encodes the global statistics of category $k$. It amortizes per-image reference vectors $\mathbf{r}$ by aggregating their information into persistent, category-level prototypes.
Given a reference vector $\mathbf{r}\in\mathbb{R}^C$ for category $y$, we update the memory with an exponential moving average (EMA):
\begin{equation}
\mathbf{m}_y \leftarrow \mu\,\mathbf{m}_y + (1-\mu)\,\mathbf{r},\quad 0<\mu<1,
\end{equation}
where $\mu$ is the momentum coefficient. At inference, given only a camouflaged query, we extract a compact descriptor $\mathbf{q}\in\mathbb{R}^C$ from the query features and predict logits $\mathbf{a}=h(\mathbf{q})\in\mathbb{R}^K$.
We then form a mixture over prototypes:
\begin{equation}
\begin{aligned}
\boldsymbol{\pi}&=\mathrm{softmax}\!\left(\mathbf{a}\right), \\
\mathbf{v}&=\sum_{k=1}^{K}\pi_k\,\mathbf{m}_k.
\end{aligned}
\end{equation}
The vector $\mathbf{v}\in\mathbb{R}^C$ serves as the reference guidance synthesized without any test-time references.

During training, when a reference of category $y$ is available, we encourage the predictor $h(\cdot)$ to produce category-consistent mixtures via a token/classification loss implemented by a standard cross-entropy loss:
\begin{equation}
\mathcal{L}_{\text{cls}}
= - \log \frac{\exp(a_y)}{\sum_{j=1}^{K} \exp(a_j)}.
\end{equation}
Optionally, to bootstrap early learning, the synthesized guidance can be combined with the current reference token, e.g., $\tilde{\mathbf{v}}=\mathbf{v}+\mathbf{r}$ when the reference is present; at inference we use $\mathbf{v}$ alone.
The memory update uses stop-gradient to avoid collapsing to trivial solutions, while the predictor $h(\cdot)$ is trained with the token loss $\mathcal{L}_{\text{cls}}$.

The above mixture-based retrieval provides a soft, query-adaptive combination of prototypes that better handles intra-class diversity than nearest-prototype selection.
Compared with test-time references, the class-prototype memory offers a stable prior with negligible overhead and no reference data collection in inference.

\begin{figure}[t]
  \centering
  \includegraphics[width=\linewidth]{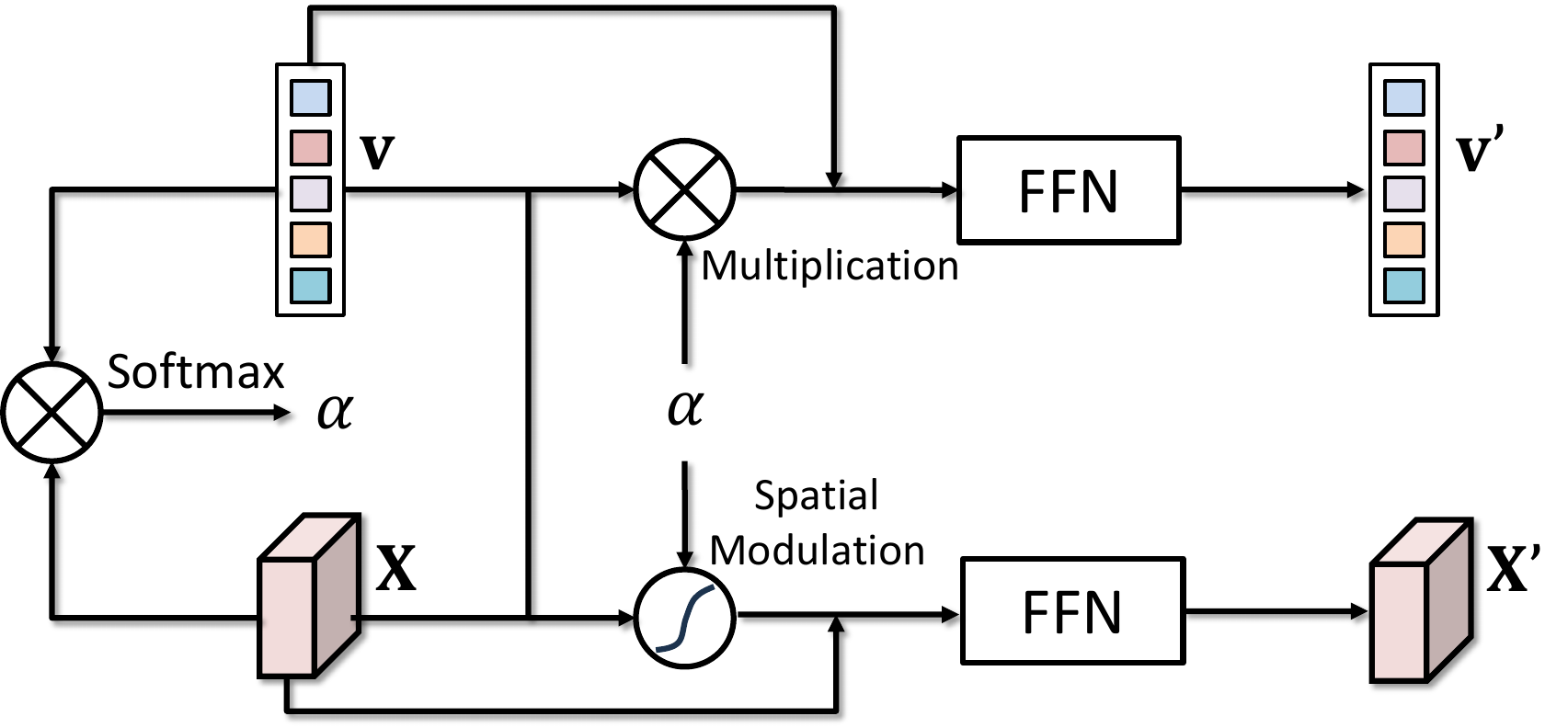}
  \vspace{-4mm}
  \caption{\textbf{Architecture of the bidirectional attention alignment.}}
  \label{fig:alignment}
\end{figure}

\subsection{Bidirectional Attention Alignment}
\label{sec:alignment}

Even with a category-aware vector synthesized from the prototype memory, effectively transferring this guidance to the camouflaged query remains a significant challenge. As highlighted in the introduction, there exists a significant gap between the high-contrast, salient reference domain and the low-contrast, cluttered camouflaged query domain. Prior methods relying on simple global affine modulation often fail because they only weakly align spatial semantics, leading to over-transfer or misalignment.
We therefore introduce a bidirectional attention alignment (BAA) mechanism that allows mutual interaction between the query features and the reference guidance, 
so that the reference vector is dynamically refined by the most discriminative query regions. 
The architecture is illustrated in \figref{fig:alignment}.

\begin{table*}[t!]
\setlength\tabcolsep{6pt}
\centering
\caption{\textbf{Comparison with recent state-of-the-art methods}. The benchmark is performed on the R2C7K test set. Methods marked with $^1$ and $^2$ use the ZoomNet \cite{pang2022zoom} and ZoomNeXt \cite{pang2024zoomnext} mask decoders, respectively.}
\vspace{-3mm}
\resizebox{\linewidth}{!}{
\begin{tabular}{llcccccccccccc}
\toprule
\multirow{2}{*}{Models} & \multirow{2}{*}{Backbone} & \multicolumn{4}{l}{\makebox[0.22\textwidth][c]{\makecell{Overall}}} & \multicolumn{4}{l}{  \makebox[0.22\textwidth][c]{\makecell{Single-obj}}} & \multicolumn{4}{l}{  \makebox[0.22\textwidth][c]{\makecell{Multi-obj}}}\\
\cmidrule(lr){3-6}
\cmidrule(lr){7-10}
\cmidrule(lr){11-14}
& & \makecell{S$_m$ $\uparrow$} &\makecell{$\alpha $E $\uparrow$}  &\makecell{$w$F  $\uparrow$} &\makecell{M$\downarrow$} & \makecell{S$_m$ $\uparrow$} &\makecell{$\alpha $E $\uparrow$}  &\makecell{$w$F  $\uparrow$} &\makecell{M$\downarrow$} & \makecell{S$_m$ $\uparrow$} &\makecell{$\alpha $E $\uparrow$}  &\makecell{$w$F  $\uparrow$} &\makecell{M$\downarrow$}\\ 
\midrule
\multicolumn{14}{l}{\textbf{CNN-based backbones}} \\
\midrule
PFNet-Ref~\cite{mei2021camouflaged}       & ResNet-50 \cite{he2016deep} & 0.811 & 0.885 & 0.687 & 0.036 & 0.815 & 0.886 & 0.691 & 0.035 & 0.764 & 0.873 & 0.632 & 0.045 \\
PreyNet-Ref~\cite{zhang2022preynet}     & ResNet-50 \cite{he2016deep} & 0.817 & 0.900 & 0.704 & 0.032 & 0.822 & 0.900 & 0.709 & 0.032 & 0.763 & \textbf{0.898} & 0.645 & 0.041 \\
SINetV2-Ref~\cite{fan2021concealed}     & Res2Net-50 \cite{gao2019res2net} & 0.823 & 0.888 & 0.700 & 0.033 & 0.828 & 0.889 & 0.705 & 0.032 & 0.771 & 0.874 & 0.634 & 0.043 \\
BSANet-Ref~\cite{zhu2022can}      & Res2Net-50 \cite{gao2019res2net}  & 0.830 & 0.912 & 0.727 & 0.030 & 0.827 & 0.913 & 0.733 & 0.030 & 0.774 & 0.895 & 0.655 & 0.039 \\
BGNet-Ref~\cite{sun2022boundary}       & Res2Net-50 \cite{gao2019res2net} & 0.840 & 0.909 & 0.738 & 0.029 & 0.844 & 0.910 & 0.742 & 0.029 & 0.792 & 0.887 & 0.679 & 0.036 \\
DGNet-Ref~\cite{ji2023deep} & EfficientNet-B4 \cite{tan2019efficientnet} & 0.821 & 0.891 & 0.696 & 0.032 & 0.827 & 0.890 & 0.703 & 0.031 & 0.748 & 0.879 & 0.607 & 0.045 \\
UAT~\cite{wu2025uncertainty} & Res2Net-50  \cite{gao2019res2net} & 0.825 & 0.886 & 0.698 & 0.033 & 0.829 & 0.884 & 0.702 & 0.032 & 0.762 & 0.862 & 0.613 & 0.044 \\
R2CNet$^1$~\cite{zhang2025referring}     &     ResNet-50  \cite{he2016deep}    & 0.834 & 0.886 & 0.720 & 0.029 & 0.839 & 0.887 & 0.726 & 0.029 & 0.781 & 0.876 & 0.652 & 0.038 \\
R2CNet$^2$~\cite{zhang2025referring}   &      ResNet-50  \cite{he2016deep}   & 0.850 & 0.909 & 0.755 & 0.027 & 0.859 & 0.910 & 0.762 & 0.026 & 0.788 & 0.892 & 0.675 & 0.037 \\
\rowcolor[HTML]{EFEFEF}
RefOnce$^1$ (Ours) &   ResNet-50   \cite{he2016deep}     & 0.846 & 0.904 & 0.743 & 0.027 & 0.851 & 0.905 & 0.749 & 0.027 & 0.799 & 0.893 & 0.684 & 0.035 \\
\rowcolor[HTML]{EFEFEF}
RefOnce$^2$ (Ours) &   ResNet-50 \cite{he2016deep} & \textbf{0.860} & \textbf{0.914} & \textbf{0.768} & \textbf{0.025} & \textbf{0.865} & \textbf{0.916} & \textbf{0.775} & \textbf{0.024} & \textbf{0.811} & 0.897 & \textbf{0.705} & \textbf{0.033} \\
\midrule
\multicolumn{14}{l}{\textbf{Transformer-based backbones}} \\
\midrule
VSCode-Ref~\cite{luo2024vscode} & Swin-S \cite{liu2021swin} & 0.832 & 0.891 & 0.714 & 0.030 & 0.838 & 0.892 & 0.718 & 0.029 & 0.766 & 0.880 & 0.662 & 0.041 \\
CIRCOD~\cite{Gupta_2025_WACV} & PVTv2-B2~\cite{wang2022pvt} & 0.848 & 0.918 & 0.756 & 0.026 & - & - & - & - & - & - & - & - \\
UAT~\cite{wu2025uncertainty} & PVTv2-B2 \cite{wang2022pvt}  & 0.855 & 0.912 & 0.757 & 0.026 & 0.859 & 0.913 & 0.761 & 0.025 & 0.805 & 0.900 & 0.701 & 0.033 \\
RPMA~\cite{liu2024reference} & SegFormer-B4 \cite{xie2021segformer}   & 0.862 & 0.930  & 0.784 & 0.023 & 0.867 & 0.934 & 0.791 & 0.023 & 0.806 & 0.894 & 0.718 & 0.033 \\
\rowcolor[HTML]{EFEFEF}
RefOnce$^2$ (Ours) &  PVTv2-B2  \cite{wang2022pvt}        & \textbf{0.890} & \textbf{0.937} & \textbf{0.819} & \textbf{0.019} & \textbf{0.894} & \textbf{0.937} & \textbf{0.825} & \textbf{0.018} & \textbf{0.853} & \textbf{0.930} & \textbf{0.767} & \textbf{0.024} \\
\bottomrule
\end{tabular}}
\vspace{1pt}
\label{tab:cmp_sota}
\end{table*}

Let $\mathbf{X}\in\mathbb{R}^{C\times H\times W}$ be a query feature map and $\mathbf{v}\in\mathbb{R}^C$ the guidance.
Flatten $\mathbf{X}$ to $\mathbf{F}\in\mathbb{R}^{HW\times C}$ and compute coupled attention scores
\begin{equation}
\begin{aligned}
s_p &= \frac{\big\langle \mathrm{LN}(\mathbf{F}_p)\mathbf{W}_x,\ \mathrm{LN}(\mathbf{v})\mathbf{W}_v\big\rangle}{\sqrt{C}},\quad p=1,\dots,HW\\
\alpha_p &= \frac{\exp(s_p)}{\sum_{t}\exp(s_t)},
\end{aligned}
\end{equation}
where LN is the LayerNorm. The spatial gate $\mathbf{G}\in\mathbb{R}^{1\times H\times W}$ is the reshaped $\alpha$.
Conditioned on $\mathbf{v}$, we generate the scaling parameters as below:
\begin{equation}
\begin{aligned}
\boldsymbol{\gamma}&=\tanh(\mathbf{W}_{\gamma}\mathbf{v}),\\
\boldsymbol{\beta}&=\tanh(\mathbf{W}_{\beta}\mathbf{v}),
\end{aligned}
\end{equation}
and apply a spatially-adaptive modulation:
\begin{equation}
\mathbf{X}'=\mathbf{X}\;+\;\mathbf{G}\odot\bigl((\mathbf{1}+\boldsymbol{\gamma})\odot \mathbf{X}+\boldsymbol{\beta}\bigr),
\end{equation}
where $\odot$ is element-wise product with appropriate broadcasting, and $\mathbf{1}$ denotes an all-ones tensor of appropriate shape.
In the reverse direction, we aggregate evidence from the image to refine the guidance:
\begin{equation}
\begin{aligned}
\mathbf{c}&=\sum_{p}\alpha_p\,\mathbf{W}_c\mathbf{F}_p, \\
\Delta\mathbf{v}&=\mathbf{W}_o\mathbf{c},\\
\mathbf{v}'&=\mathbf{v}+\Delta\mathbf{v}.
\end{aligned}
\end{equation}
We further stabilize both branches with residual feed-forward updates:
\begin{equation}
\begin{aligned}
\mathbf{v}'&\leftarrow \mathbf{v}'+\mathrm{FFN}\!\bigl(\mathrm{LN}(\mathbf{v}')\bigr)\\
\mathbf{X}'&\leftarrow \mathbf{X}'+\mathrm{FFN}\!\bigl(\mathrm{LN}(\mathbf{X}')\bigr)
\end{aligned}
,
\end{equation}
where FFN is the feed-forward network with two linear layers and GELU \cite{hendrycks2016gaussian} activation. $\mathbf{v}'$ and $\mathbf{X}'$ are refined reference vector and camouflaged features, respectively. 

The whole process is iterable and we can run the above process for multiple times for better performance. 
Reference features are 1-d vectors which provide object-level guidance to camouflaged features, so we choose to run the above process three times on the high-level backbone features $\mathbf{X_4}$ 
while we only run once for other stage backbone features with the refined $\mathbf{v'}$ as the reference.

\section{Experiments}

\subsection{Setup}

\para{Datasets.}
We conduct our experiments primarily on the R2C7K dataset, a newly introduced benchmark comprising 6,615 samples across 64 categories. 
Each category contains 25 reference images, resulting in a total of 1,600 reference images. 
By default, we train each network on the training set of the R2C7K dataset.
To further evaluate the generalization ability of our approach to conventional camouflage object detection (COD) scenarios, we also assess its performance on four widely used COD datasets: CAMO \cite{le2019anabranch}, CHAMELEON \cite{skurowski2018animal}, COD10K \cite{fan2020camouflaged}, and NC4K \cite{lv2021simultaneously}. 
Moreover, to rigorously test cross-category generalization, we curate an Unseen Classes dataset by selecting 596 images from these four datasets whose categories are not included among the 64 categories in R2C7K.

\para{Evaluation metrics.}
We perform the benchmarks under the standard Ref-COD protocol. We report mean absolute error (MAE), structure-measure \cite{fan2017structure} (S$_m$), adaptive E-measure \cite{fan2018enhanced} ($\alpha$E), and weighted F-measure \cite{margolin2014evaluate} ($w$F): MAE (M) measures absolute difference to the ground truth; S$_m$ captures region- and object-aware structural similarity; $\alpha$E combines element-wise and image-level similarity; and $w$F integrates precision and recall with spatial weighting.

\para{Implementation details.}
We implement our RefOnce via the PyTorch \cite{paszke2019pytorch} framework. During training, we use the SGD (momentum 0.9, weight decay $5\times10^{-4}$) optimizer with initial learning rate of 0.05, linear warming up, and linear decay strategy. We train the network for 40 epochs with a batch size of 8. All images are resized to $384\times384$. During training, we apply random crop, resize, and rotation as data augmentation. For each camouflaged query, we follow previous works \cite{zhang2025referring, wu2025uncertainty} to sample five same-category reference images and average the computed reference features into a single vector to condition the network during training. 
During inference, the network only relies on the class-prototype memory for camouflaged object references. 
The predictor $h(\cdot)$ is implemented by 2 linear layers and ReLU activation.

\subsection{Results}

\def\predHeader{%
1) Input Image, 2) Ref Image, 3) GT, 4) ZoomNet, 5) R2CNet (ZoomNet) \cite{zhang2025referring}, 6) Ours (ZoomNet)%
}
\def\predPrefixList{%
COD10K-CAM-2-Terrestrial-24-Caterpillar-1594,
COD10K-CAM-2-Terrestrial-34-Human-1991,
COD10K-CAM-3-Flying-55-Butterfly-3418,
COD10K-CAM-3-Flying-61-Katydid-4004%
}
\def\predSuffixList{%
.jpg,
_a_ref_img,
.png,
_zoomnet,
_zoomnet_ref,
_a%
}

\newcommand{\AddImg}[1]{%
    \includegraphics[height=0.075\textwidth]{imgs/prediction/COD10K-CAM-2-Terrestrial-24-Caterpillar-1594#1} &%
    \includegraphics[height=0.075\textwidth]{imgs/prediction/COD10K-CAM-2-Terrestrial-34-Human-1991#1} &%
    \includegraphics[height=0.075\textwidth]{imgs/prediction/COD10K-CAM-3-Flying-55-Butterfly-3418#1} &%
    \includegraphics[height=0.075\textwidth]{imgs/prediction/COD10K-CAM-3-Flying-61-Katydid-4004#1} %
}

\begin{figure}[!t]
    \centering
    \footnotesize
    \renewcommand{\arraystretch}{0.1}
    \setlength{\tabcolsep}{0.3mm}
    \resizebox{\linewidth}{!}{%
    \begin{tabular}{ccccc}
        \rotatebox[origin=l]{90}{\hspace{-1mm} Input Image} & \AddImg{.jpg} \\ \\[0.1mm]
        \rotatebox[origin=l]{90}{Ref Image}& \AddImg{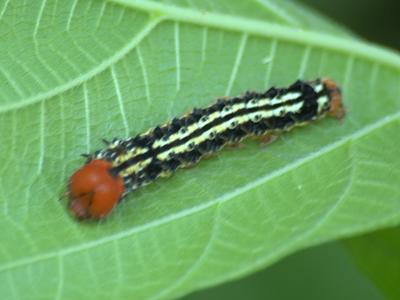} \\ \\[0.1mm]
        \rotatebox[origin=l]{90}{\hspace{2mm} GT} & \AddImg{.png} \\ \\[0.1mm]
        \rotatebox[origin=l]{90}{ZoomNet} & \AddImg{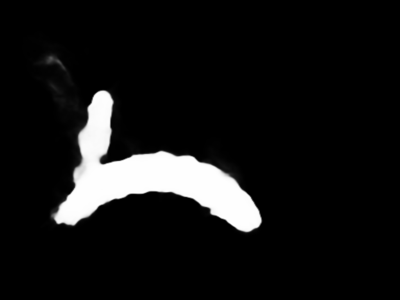} \\ \\[0.1mm]
        \rotatebox[origin=l]{90}{R2CNet} & \AddImg{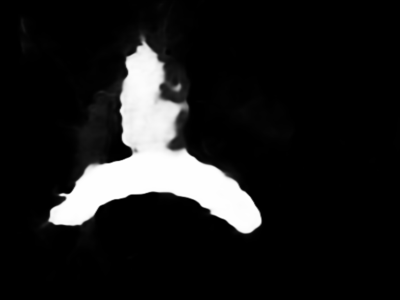} \\ \\[0.1mm]
        \rotatebox[origin=l]{90}{\hspace{2mm} Ours} & \AddImg{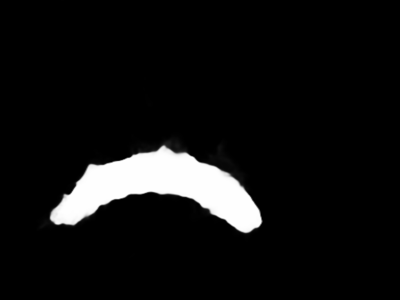} \\ \\[0.1mm]
    \end{tabular}
    }
    \vspace{-3mm}
    \caption{\textbf{Qualitative comparison on R2C7K dataset}.
    }
    \label{fig:cmp_vis}
\end{figure}%

\def\visualHeader{%
1) Input Image, 2) Ref Image, 3) GT, 4) w/o Ref, 5) w/ R2C \cite{zhang2025referring}, 6) w/ BAA (Ours)%
}
\def\visualPrefixList{%
COD10K-CAM-1-Aquatic-2-ClownFish-15,
COD10K-CAM-1-Aquatic-6-Fish-150,
COD10K-CAM-2-Terrestrial-23-Cat-1392,
COD10K-CAM-2-Terrestrial-45-Spider-2614,
COD10K-CAM-3-Flying-53-Bird-3223%
}
\def\visualSuffixList{%
_a_1,
_a_ref_img,
_a_2,
_d_1_bb,
_c_zn_ref,
_d_2_ref%
}

\begin{figure*}[t]
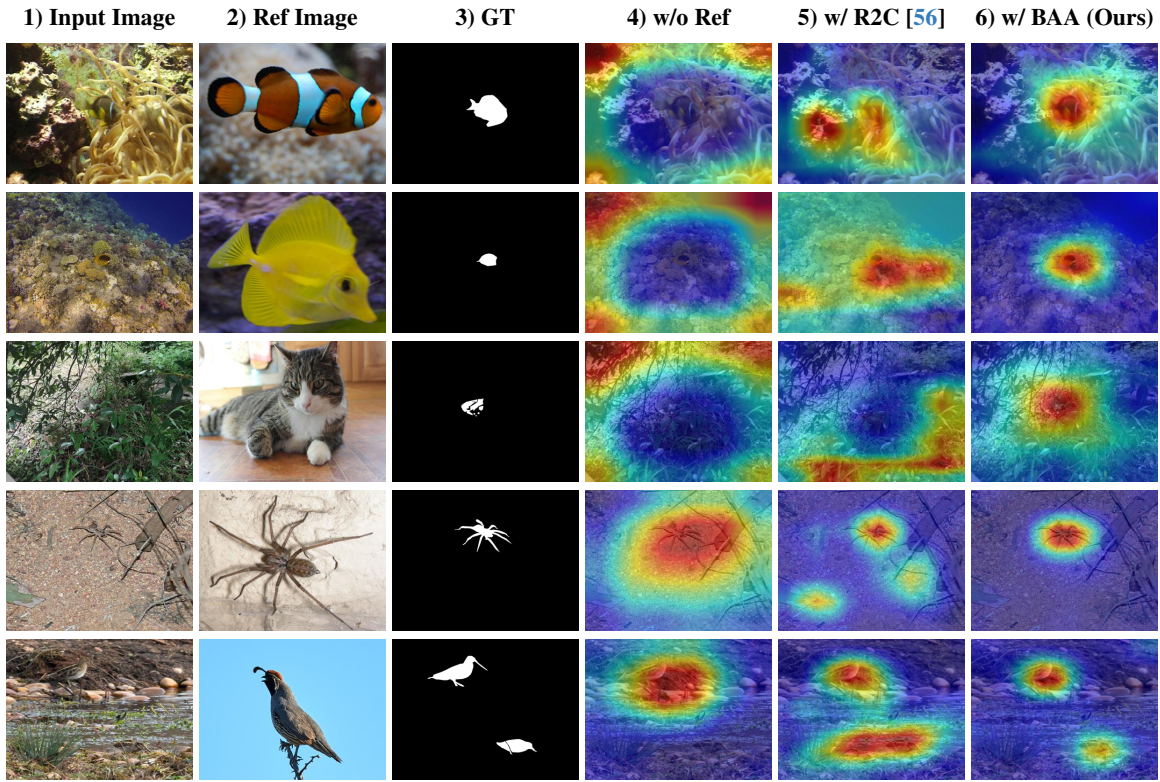

  \centering
  \imagegrid{imgs/visualization/}{\visualHeader}{\visualPrefixList}{\visualSuffixList}{0.142\linewidth}{\small\textbf}
  \vspace{-3mm}
  \caption{\textbf{Visualizations of backbone features with or without reference guidance}. ``4) w/o Ref'' indicates the visualization of backbone features before reference guidance. 
  ``5) w/ R2C'' is the visualization of backbone features with R2C \cite{zhang2025referring} as reference guidance.
  ``6) w/ RefOnce (Ours)'' replaces R2C with our BAA (\secref{sec:alignment}) strategy.
  More examples can refer to the supplementary.}
  \label{fig:visualization}
\end{figure*}

\para{Comparison with other state-of-the-arts.}
As shown in \tabref{tab:cmp_sota}, our RefOnce consistently surpasses all recent Ref-COD approaches under both CNN and transformer backbones.
For CNN-based settings, RefOnce (ResNet-50) already outperforms representative reference-dependent models such as R2CNet \cite{zhang2025referring}, UAT \cite{wu2025uncertainty}, and RPMA \cite{liu2024reference}, achieving higher accuracy without using any reference images at inference.
Notably, RefOnce (PVTv2-B2) obtains the best performance with $\text{S}_m=0.890$ and $\alpha\text{E}=0.937$, establishing a new state-of-the-art on the R2C7K benchmark.
These results demonstrate that our prototype memory and BAA effectively retain the semantic benefits of reference guidance while removing its inference-time dependency.

\para{Qualitative comparison.}
\figref{fig:cmp_vis} presents qualitative comparisons on R2C7K, where RefOnce produces more complete and precise object boundaries than prior methods. The masks generated by RefOnce are cleaner and better aligned with the concealed targets, especially in challenging low-light or cluttered scenes. For example, in the second example from the left, the human in a dark environment is barely detected by other models, while our approach correctly segments the entire figure. These visual results confirm that RefOnce yields more reliable and fine-grained detection even without any test-time references.

\subsection{Ablation Study}

\para{Significance of reference for COD.}
In this part, we analyze how reference information helps the network recognize camouflaged objects. 
COD is an inherently challenging task because the target often blends almost perfectly into the background, causing the backbone network to fail in capturing object-level semantics.
As shown in \figref{fig:visualization}, without any reference guidance (w/o Ref, column 4), the model’s activation maps are scattered and mainly respond to background textures, indicating poor localization of the camouflaged object. Incorporating reference information provides strong semantic cues that help the model focus on the correct regions.
The basic Ref-COD framework, R2C \cite{zhang2025referring} (w/ R2C, column 5), partially alleviates this issue by introducing external guidance, yet the response remains inaccurate or incomplete due to weak feature alignment between reference and query. 
In contrast, our BAA-based approach (w/ BAA (Ours), column 6) effectively aligns the synthesized reference representation with the query features, enabling precise localization of the concealed target. 
Importantly, our model achieves this without requiring any reference image during inference, demonstrating that the distilled prototype memory and bidirectional alignment module allow the network to internalize reference cues and utilize them implicitly for more accurate and robust detection.

\begin{table}[t!]
\setlength\tabcolsep{3pt}
\centering
\caption{\textbf{Evaluation of the generalization ability of our RefOnce}. The baseline is our method training without any reference guidance. Note that the performance gain ($\Delta$) is most pronounced on the Unseen Classes subset, highlighting our model's strong generalization to novel categories.}
\vspace{-3mm}
\resizebox{\linewidth}{!}{
\begin{tabular}{llcccc}
\toprule
Dataset & Method & \makecell{S$_m$ $\uparrow$} & \makecell{$\alpha$E $\uparrow$} & \makecell{$w$F $\uparrow$} & \makecell{M $\downarrow$} \\
\midrule
\multirow{3}{*}{\makecell{CHAMELEON}} 
& Baseline (w/o Ref) & 0.853 & 0.938 & 0.773 & 0.034 \\
& Ours & 0.882 & 0.941 & 0.812 & 0.029 \\
\cline{2-6}
& \textit{$\Delta$} & \textit{+0.029} & \textit{+0.003} & \textit{+0.039} & \textit{-0.005} \\
\midrule
\multirow{3}{*}{\makecell{CAMO}} 
& Baseline (w/o Ref) & 0.698 & 0.794 & 0.564 & 0.108 \\
& Ours & 0.728 & 0.801 & 0.615 & 0.098 \\
\cline{2-6}
& \textit{$\Delta$} & \textit{+0.030} & \textit{+0.007} & \textit{+0.051} & \textit{-0.010} \\
\midrule
\multirow{3}{*}{\makecell{COD10K}} 
& Baseline (w/o Ref) & 0.817 & 0.881 & 0.695 & 0.033 \\
& Ours & 0.844 & 0.902 & 0.741 & 0.028 \\
\cline{2-6}
& \textit{$\Delta$} & \textit{+0.027} & \textit{+0.021} & \textit{+0.046} & \textit{-0.005} \\
\midrule
\multirow{3}{*}{\makecell{NC4K}} 
& Baseline (w/o Ref) & 0.824 & 0.890 & 0.744 & 0.052 \\
& Ours & 0.854 & 0.910 & 0.789 & 0.043 \\
\cline{2-6}
& \textit{$\Delta$} & \textit{+0.030} & \textit{+0.020} & \textit{+0.045} & \textit{-0.009} \\
\midrule
\multirow{3}{*}{\makecell{Unseen Classes}} 
& Baseline (w/o Ref) & 0.801 & 0.868 & 0.713 & 0.061 \\
& Ours & 0.839 & 0.892 & 0.770 & 0.048 \\
\cline{2-6}
& \textit{$\Delta$} & \textit{+0.038} & \textit{+0.024} & \textit{+0.057} & \textit{-0.013} \\
\bottomrule
\end{tabular}}
\vspace{2pt}
\label{tab:generalization}
\end{table}

\para{Generalization ability.}
To evaluate the generalization ability of our framework, we compare it with the baseline that removes the reference branch and trains the network without any reference guidance, while keeping other settings identical. 
Both the baseline and our method are trained on the R2C7K dataset.
Results are shown in \tabref{tab:generalization}, which includes four standard COD benchmarks and one additional Unseen Classes subset. The Unseen Classes dataset is constructed from the above four COD benchmarks but deliberately excludes all categories contained in all classes of R2C7K.
This setting aims to test the model’s ability to generalize to novel categories never observed during training.
We can observe that our model surpasses the baseline by a large margin in all datasets.
Notably, the performance gain is most pronounced on the Unseen Classes subset (an S$_m$ gain of +0.038, compared to $\sim$+0.027-0.030 on other datasets).
This key finding highlights the baseline's vulnerability to novel categories, where its performance is weakest due to the lack of any relevant priors.
In contrast, our model, by synthesizing a transferable semantic prior via the prototype mixture mechanism, provides the most significant relative advantage in this challenging scenario.
This strongly suggests that the proposed prototype memory effectively captures transferable semantic priors, not just memorizing the training categories.
This capability, combined with our model's reference-free inference, allows RefOnce to generalize smoothly to new categories, a task where previous reference-dependent methods fundamentally fail.

\para{Momentum updates.}
As shown in \tabref{tab:abl-momentum}, varying the momentum coefficient $\mu$ used for updating the prototype memory slightly affects the performance. The model achieves the best overall balance when $\mu=0.99$, indicating that a relatively slow update helps maintain stable category representations while still allowing gradual adaptation. Therefore, we adopt $\mu=0.99$ as the default setting.

\para{Loss factor.}
As presented in \tabref{tab:abl-loss-cls}, the weight $\lambda_c$ of the classification loss significantly affects the performance.
A moderate value of $\lambda_c$ yields the best results, suggesting that an appropriate supervision strength effectively guides the mixture predictor without over-constraining it.
Based on the results, we choose $\lambda_c=0.03$ as the default setting.

\begin{table}[tp]
  \centering
  \small
  \caption{\textbf{Ablation study on the loss factor $\lambda_c$ of $\mathcal{L}_{\text{cls}}$ in generating the reference vector $\mathbf{v}$}.}
  \vspace{-3mm}
  \setlength\tabcolsep{10pt}
  \begin{tabular}{llcccccc} \toprule
    No. & $\lambda_c$ & S$_m \uparrow$ & $\alpha $E $\uparrow$  & $w$F $\uparrow$ & M$\downarrow$ \\ \midrule
    1 & 0.3   & 0.794 & 0.856 & 0.650 & 0.037 \\
    2 & 0.1   & 0.830 & 0.886 & 0.713 & 0.030 \\
    \rowcolor[HTML]{EFEFEF}
    3 & 0.03  & \textbf{0.846} & \textbf{0.904} & \textbf{0.743} & \textbf{0.027} \\
    4 & 0.01  & 0.838 & 0.893 & 0.730 & 0.029 \\
    5 & 0.003 & 0.835 & 0.896 & 0.730 & 0.029 \\
    \bottomrule
  \end{tabular}
  \label{tab:abl-loss-cls}
\end{table}

\begin{table}[tp]
  \centering
  \small
  \caption{\textbf{Ablation study on the momentum $\mu$ of updating the memory $\mathcal{M}$}.}
  \vspace{-3mm}
  \setlength\tabcolsep{10pt}
  \begin{tabular}{llcccccc} \toprule
    No. & $\mu$ & S$_m \uparrow$ & $\alpha $E $\uparrow$  & $w$F $\uparrow$ & M$\downarrow$ \\ \midrule
    1 & 0.9   & 0.844 & \textbf{0.904} & 0.742 & 0.028 \\
    \rowcolor[HTML]{EFEFEF}
    2 & 0.99  & \textbf{0.846} & \textbf{0.904} & \textbf{0.743} & \textbf{0.027} \\
    3 & 0.999  & 0.844 & 0.901 & 0.739 & \textbf{0.027} \\
    \bottomrule
  \end{tabular}
  \label{tab:abl-momentum}
\end{table}

\para{Mixture over prototypes.}
To investigate the effect of synthesizing the reference vector, we compare two strategies: (1) mixture over prototypes, which forms a soft combination of all category prototypes using query-conditioned mixture weights; 
(2) select the nearest one, which simply selects the prototype corresponding to the largest predicted logit; 
(3) use the GT class, an upper-bound case where the ground-truth category of each query is provided during inference.
As shown in \tabref{tab:abl-mixture}, the mixture-based strategy consistently improves all metrics over the nearest-prototype selection and approaches the upper-bound performance.
These results demonstrate that soft combination over prototypes yields a more stable and informative reference representation, effectively approximating the ideal class-specific guidance without requiring any ground-truth information at test time.

\para{BAA.}
Here we compare the BAA with the traditional one-way global modulation, which removes the inverse direction of refining the reference features in BAA.
We observe that removing this direction in BAA results in a degradation of 0.4\% S$_m$, 0.9\% $\alpha$E, 1.0\% $w$F, and 0.1\% MAE.
This confirms that the reverse alignment, which allows the guidance vector to be refined by image-specific features, is crucial for effective feature adaptation and bridging the domain gap.

\para{Computational overhead.}
We analyze the computational cost of our RefOnce, as shown in \tabref{tab:params_flops}.
We report the FLOPs and parameter counts of each major component, including reference feature extraction, reference generation, reference guidance, and the remaining parts (camouflaged feature extraction and mask prediction).
During inference, the reference feature extraction module is omitted since RefOnce is reference-free.
As shown in the table, this module accounts for a substantial portion of the total cost, comparable to the rest of the network.
In contrast, reference generation is lightweight, requiring only 0.01 GFLOPs, and the reference guidance introduces an 11.8\% overhead compared with the whole model during inference.

\begin{table}[tp]
  \centering
  \small
  \caption{\textbf{Effect of mixture over prototypes}. We additionally report the upper-bound performance when the ground-truth class is provided during inference (Use the GT Class).}
  \vspace{-3mm}
  \setlength\tabcolsep{6pt}
  \begin{tabular}{llcccccc} \toprule
    Setting & S$_m \uparrow$ & $\alpha $E $\uparrow$  & $w$F $\uparrow$ & M$\downarrow$ \\ \midrule
    Select the Nearest One   & 0.840 & 0.894 & 0.731 & 0.028 \\
    \rowcolor[HTML]{EFEFEF}
    Mixture over Prototypes  & 0.846 & 0.904 & 0.743 & 0.027 \\
    Oracle: Use the GT Class & 0.851  &  0.904  &  0.751  & 0.026 \\
    \bottomrule
  \end{tabular}
  \label{tab:abl-mixture}
\end{table}

\begin{table}[tp]
  \centering
  \small
  \caption{\textbf{Computational cost analysis of RefOnce}. \textcolor{gray}{Gray} texts indicate modules that are omitted during inference.}
  \vspace{-3mm}
  \setlength\tabcolsep{8pt}
  \begin{tabular}{lcc} \toprule
    Module & \# Params (M) & FLOPs (G) \\ \midrule
    \textcolor{gray}
    {Reference Feature Extraction}        & \textcolor{gray}{33.1} & \textcolor{gray}{124.5}  \\
    Reference Generation & 0.1 & 0.01  \\
    Reference Guidance & 2.5 & 13.7 \\ 
    Other parts & 32.5 & 102.0   \\
    \bottomrule
  \end{tabular}
  \label{tab:params_flops}
\end{table} 

\section{Conclusion}

In this work, we introduced \textbf{RefOnce}, a reference-free framework for referring camouflaged object detection (Ref-COD). 
Unlike existing methods that require reference images during inference, RefOnce distills category-level semantics into a class-prototype memory and synthesizes query-conditioned guidance via a mixture over prototypes. 
A bidirectional attention alignment module further bridges the salient-camouflage domain gap by jointly adapting query features and class representations. 
Extensive experiments on the large-scale R2C7K benchmark and multiple standard COD datasets demonstrate that our approach achieves new state-of-the-art performance and exhibits strong generalization to unseen categories, even without any test-time references. 
We believe RefOnce provides a promising step toward deployable reference-guided segmentation without dependency on external examples.

{
    \small
    \bibliographystyle{ieeenat_fullname}
    \bibliography{main}
}

\end{document}

%% file: utils.tex
\usepackage{graphicx}
\usepackage{pgffor}
\usepackage{xparse}
\usepackage{array}

\newcommand{\gridcolsep}{\hspace{0.005\linewidth}}%
\newcommand{\gridrowsep}{\vspace{0.005\linewidth}}%
\newcommand{\headerfmt}[2]{#1{#2}}

\NewDocumentCommand{\imagegrid}{m m m m m m}{%
  \begingroup
  \foreach \s [count=\colidx from 1] in #2 {%
    \ifnum\colidx>1\gridcolsep\fi
    \begin{subfigure}[t]{#5}
        \parbox[t]{\linewidth}{\centering \headerfmt{#6}{\s}}
    \end{subfigure}%
  }%
  \par\vspace{0.4em}%
  \def\folder{#1}%
  \foreach \p [count=\rowidx from 1] in #3 {%
    \foreach \s [count=\colidx from 1] in #4 {%
      \ifnum\colidx>1\gridcolsep\fi
      \begin{subfigure}[t]{#5}
        \centering
        \edef\basename{\folder\p\s}%
        \typeout{\basename}
        \includegraphics[width=\linewidth, height=0.75\linewidth]{\basename}%
      \end{subfigure}%
    }%
    \par\gridrowsep%
  }%
  \endgroup
}

\NewDocumentCommand{\imagegridV}{m m m m m m}{%
  \begingroup
  \foreach \s [count=\colidx from 1] in #2 {%
    \ifnum\colidx>1\gridcolsep\fi
    \begin{subfigure}[t]{#5}
        \parbox[t]{\linewidth}{\centering \headerfmt{#6}{\s}}
    \end{subfigure}%
  }%
  \par\vspace{0.4em}%
  \def\folder{#1}%
  \foreach \p [count=\rowidx from 1] in #4 {%
    \foreach \s [count=\colidx from 1] in #3 {%
      \ifnum\colidx>1\gridcolsep\fi
      \begin{subfigure}[t]{#5}
        \centering
        \edef\basename{\folder\p\s}%
        \typeout{\basename}
        \includegraphics[width=\linewidth, height=0.75\linewidth]{\basename}%
      \end{subfigure}%
    }%
    \par\gridrowsep%
  }%
  \endgroup
}